\documentclass[conference]{IEEEtran}
\IEEEoverridecommandlockouts
\usepackage{cite}
\usepackage{amsmath,amssymb,amsfonts}
\usepackage{array}

\usepackage[bookmarks=false]{hyperref}
\hypersetup{
    colorlinks=true,
    linkcolor=blue,
    filecolor=blue,      
    urlcolor=blue,
}

\usepackage[ruled,vlined]{algorithm2e}
\usepackage{graphicx}
\usepackage{textcomp}
\usepackage{xcolor}
\def\BibTeX{{\rm B\kern-.05em{\sc i\kern-.025em b}\kern-.08em
    T\kern-.1667em\lower.7ex\hbox{E}\kern-.125emX}}

\usepackage{tikz}
\usepackage{textcomp}
\usepackage{hyperref}
\usepackage{lipsum}

\newcommand\copyrighttext{%
  \footnotesize \textcopyright 2020 IEEE.  Personal use of this material is permitted. Permission from IEEE must be obtained for all other uses, in any current or future media, including reprinting/republishing this material for advertising or promotional purposes, creating new collective works, for resale or redistribution to servers or lists, or reuse of any copyrighted component of this work in other works.}
\newcommand\copyrightnotice{%
\begin{tikzpicture}[remember picture,overlay]
\node[anchor=south,yshift=10pt] at (current page.south) {\fbox{\parbox{\dimexpr\textwidth-\fboxsep-\fboxrule\relax}{\copyrighttext}}};
\end{tikzpicture}%
}

\begin{document}

\title{A Metric Learning Approach to Anomaly Detection in Video Games}

\author{\IEEEauthorblockN{Benedict Wilkins}
\IEEEauthorblockA{\textit{Department of Computer Science} \\
\textit{Royal Holloway University of London}\\
London, England \\
Benedict.Wilkins.2014@rhul.ac.uk}
\and
\IEEEauthorblockN{Chris Watkins}
\IEEEauthorblockA{\textit{Department of Computer Science} \\
\textit{Royal Holloway University of London}\\
London, England \\
C.J.Watkins@rhul.ac.uk}
\and
\IEEEauthorblockN{Kostas Stathis}
\IEEEauthorblockA{\textit{Department of Computer Science} \\
\textit{Royal Holloway University of London}\\
London, England \\
Kostas.Stathis@rhul.ac.uk}
}

\maketitle
\copyrightnotice

\begin{abstract}
With the aim of designing automated tools that assist in the video game quality assurance process, we frame the problem of identifying bugs in video games as an anomaly detection (AD) problem. We develop State-State Siamese Networks (S3N) as an efficient deep metric learning approach to AD in this context and explore how it may be used as part of an automated testing tool. Finally, we show by empirical evaluation on a series of Atari games, that S3N is able to learn a meaningful embedding, and consequently is able to identify various common types of video game bugs.

\end{abstract}


\newcommand{\gamegraph}{game graph} 
\newcommand{\truegame}{true game} 
\newcommand{\falsegame}{false game} 

\section{Introduction}


Video game development companies take significant steps at all stages of development to reduce the likelihood of bugs appearing in release code. These steps range from the use of software development paradigms early in the process to heavy investment in Quality Assurance (QA) closer to release. As games become increasingly vast and complex, exploring and uncovering bugs manually is becoming less feasible \cite{Chang2019}. In contrast, the continuing advancements in Reinforcement Learning (RL) are allowing software agents to play and explore with greater proficiency in increasingly complex games. This has opened up an opportunity for the development of automated tools to assist developers and testers in the video game QA process. Previous attempts in developing these tools have focused on building frameworks\cite{Nantes2008} or require detailed descriptions of the environment and are heavily integrated with the games internal implementation \cite{Nantes2013}. 

With the aim of developing automated testing tools that can be easily integrated with existing development practices, we frame the problem of identifying bugs as an Anomaly Detection (AD) problem, treating the manifestation of a bug in the raw observation space (as seen by a human player) as an anomaly. With this view, we explore deep metric learning as an approach to AD, and its potential to form the basis for such tools. 

Specifically, we formalise the AD problem in this context and present State-State Siamese Networks (S3N) as a semi-supervised metric learning approach. S3N uses spatial and local temporal information to efficiently learn a latent representation of the state space that induces a meaningful measure of normality. We use Atari games from the Arcade Learning Environment (ALE) \cite{Bellemare2013} to create an open dataset of anomalies, the Atari Anomaly Dataset (AAD). The dataset consists of trajectories from 7 Atari games collected using model-free RL with common types of bugs \cite{Lewis2010} introduced artificially. Finally, we evaluate S3N's ability to construct meaningful representations and consequently its ability to detect anomalies on AAD, and discuss promising future directions.

\begin{figure}
    \centering
    \includegraphics[width=.75\linewidth, trim={0.4cm .5cm 0 .5cm},clip]{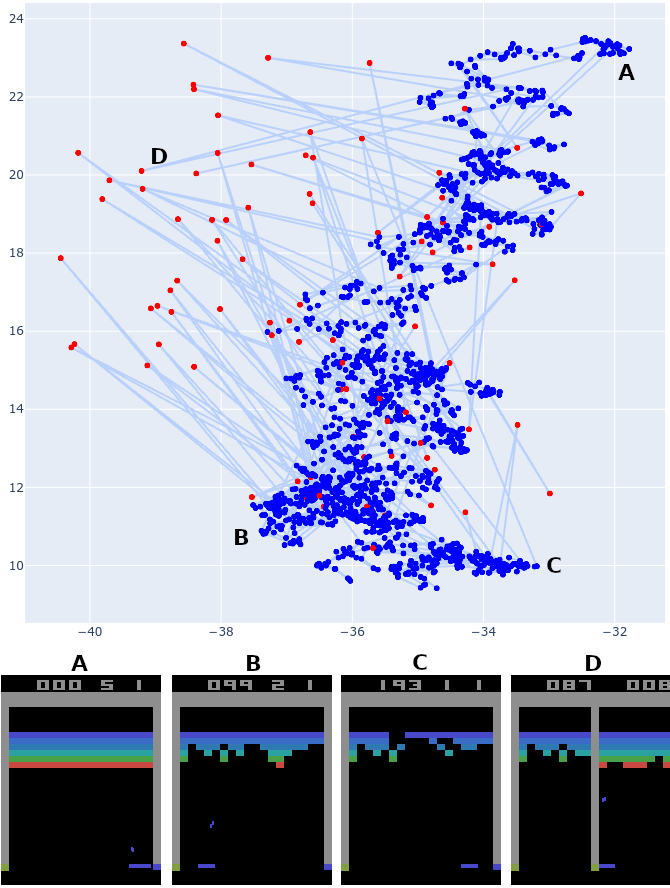}
    \caption{A purely illustrative 2D embedding space of the Atari game Breakout for a single trajectory. States are represented as points (blue=normal, red=anomalous). The distance between consecutive states is used directly as an anomaly score. Red points within the cluster of blue points resemble normal states, but are in fact anomalous with respect to the associated transitions. Crucially they are distant from their immediate transition neighbours.}
    \label{fig:latent2D}
\end{figure}
 
\section{Background \& Related Work}

\newcommand{\states}{\mathcal{S}}
\newcommand{\actions}{\mathcal{A}}
\newcommand{\tran}{\mathcal{T}}

\subsection{Formalism}

We use the following formalism for the remainder of the paper. We refer to a single frame (image) of a video game at time $t$ as a state $s_t \in \states$. A player action $a_t \in \actions$ leads to a (stochastic) transition from the state $s_t$ to $s_{t+1}$ according to a transition function $\tran(S_{t+1}|A_t, S_t, \cdots A_0, S_0)$. To simplify our discussion we consider a Markovian transition function $\tran(S_{t+1}|A_t, S_t)$. We refer to a single play-through of a game (from an initial state to a final state) as a \textit{trajectory} $\tau$. Under this formalism, a game is modelled as a labelled directed graph $(\states, \actions, \tran)$, with nodes as states and edges as transitions with associated action labels and probabilities. The development process (including QA) can be thought of as the incremental improvement of successive graphs that are \textit{closer} to some ideal graph. The graph that is released to customers being the closest approximation to the ideal graph. We denote the ideal graph as $(\states, \actions, \tran)$ and an approximate graph as $(\hat\states, \hat\actions, \hat\tran)$. We assume no prior knowledge of the state space or transition probabilities, and a small constant frame-rate. 


\subsection{Anomaly Detection (AD) in Video Games}
As it is common in video games for particular states or transitions to be rare, we cannot take the view of \textit{anomalies as outliers}. It is also common for games to have a large branching factor, in the worst cases states that occur later in time are exponentially more unlikely than their predecessors. With this in mind, we take the \textit{out of distribution} view, and define two types of anomaly:
\begin{enumerate}
    \item \textbf{State anomaly} - a state $s \in \hat\states$ is anomalous iff $s \not\in \states$
    \item \textbf{Transition anomaly} - a transition $s_{t+1} \sim \hat\tran$ is anomalous iff $s_{t+1} \not\in supp(\tran)$.
\end{enumerate}

\subsection{Siamese Networks}


Siamese networks are a general approach to metric learning, and have been successfully applied to many areas, most notably for learning image similarities in facial recognition \cite{Schroff2015}. Siamese networks learn an implicit distance by learning to represent examples in low dimensional space according to a distance-based objective \cite{Chopra2005}. They are trained on pairs of examples $(x_i, x_j)$, requiring some labelling that is indicative of the desired latent structure. 
One popular distance-based objective is triplet loss \cite{Schroff2015}:
\begin{align}
    L(x,x^+,x^-) = max(d(x, x^+) - d(x, x^-) + \alpha, 0) \\
    d(x,y) = || f_\theta(x) - f_\theta(y) || \nonumber
\end{align} where $f_\theta$ is typically a neural network with parameters $\theta$, $x$ is an anchor example, $x^+$ is an example with the same label as $x$ and $x^-$ is an example with a label that differs from $x$. Triplet loss is derived from the desired property $|| f(x) - f(x^+) || \leq || f(x) - f(x^-) ||$. The margin parameter $\alpha$ prevents the network from learning trivial solutions. Many other objectives exist \cite{Chopra2005, Chechik2010}, triplet loss is the objective that is used in our experiments. 

More recently, metric learning and specifically siamese networks, have been applied to anomaly detection \cite{Masana2018}. The key idea is that instead of using a \textit{proxy} anomaly score (e.g. reconstruction error), the score is learnt directly. The anomaly score is used to rank examples by their normality, with higher scores typically indicating abnormality. 


\begin{figure*}
    \centering
     \begin{tabular}{ccccc}
        \includegraphics[trim=10 5 10 10,clip,height=.15\linewidth]{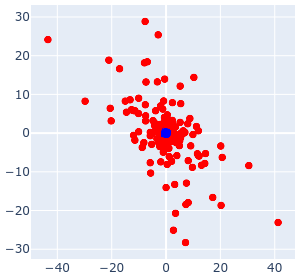} &
        \includegraphics[trim=10 0 10 0,clip,height=.15\linewidth]{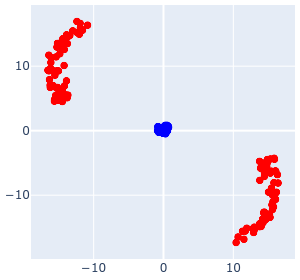} &
        \includegraphics[trim=10 0 10 0,clip,height=.15\linewidth]{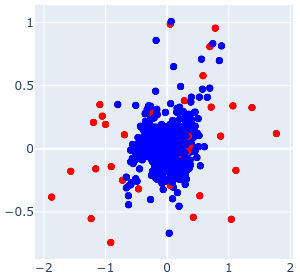} &
        \includegraphics[trim=10 0 10 0,clip,height=.15\linewidth]{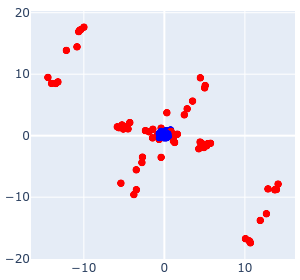} &
        \includegraphics[trim=10 0 0 0,clip,height=.15\linewidth]{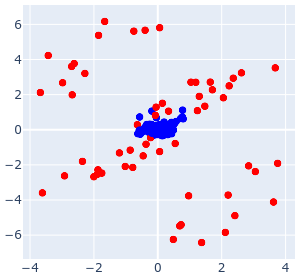} \\
        \footnotesize a) Visual artefact & \footnotesize b) Flicker & \footnotesize c) Freeze skip & \footnotesize d) Split horizontal & \footnotesize e) Split vertical \\
    \end{tabular}
    \caption{Illustrative plots of distance vectors $f_\theta(s_t) - f_\theta(s_{t+1})$ for a 2D embedding of Breakout. Blue and red points correspond to normal and anomalous transitions respectively.}
    \label{fig:distance}
\end{figure*}

\section{State-State Siamese Networks (S3N)}
S3N is a data-efficient learning procedure that is able to construct meaningful embeddings without the use of action information or a direct labelling of normal/anomalous states or transitions. S3N consists of a dynamic labelling schema and training procedure, the labelling schema is given below:
\begin{align*}
    x\ \ & = s_t  \\
    x^+ & = s_{t+1} \\
    x^- & = s_{i}, \ \ i \not\in \{t, t+1\}
\end{align*}
Under this labelling schema, states that have a temporal relationship are considered close according to the learned metric. That is, the network will attempt to embed the game graph, with connected nodes mapped to similar regions of the embedding space. We hope then, that the support of $\tran$ is in some sense captured by the neighbourhood of the particular node $s_t$ in the embedding. The desired property is given below:
\begin{align*}
    || f_\theta(s_t) - f_\theta(s_{t+1}) || \leq || f_\theta(s_t) - f_\theta(s_{i}) ||
\end{align*}
In later discussion we refer to $\Delta^j_k(\tau) = || f_\theta(s_j) - f_\theta(s_k) ||$ as the displacement with reference to a particular trajectory $\tau$. We do not impose any additional constraints on the embedding structure, and have found in our experiments that the embedding is meaningful with respect to the AD problem, see Fig. \ref{fig:latent2D}.
The learned metric evaluated on a particular query pair $(s_j, s_k)$, $\Delta^j_k$ can be used directly as an anomaly score, with anomalous transitions indicated by high $\Delta^t_{t+1}$, or low $\Delta^t_i$, see Fig. \ref{fig:distance}.


\SetKwBlock{Repeat}{Repeat:}{until}

\begin{algorithm}
    \textbf{Input:} batch size $n$; margin $\alpha$; trajectory collector $D_\pi$; neural network $f_\theta$; learning rate $\eta$. \\
    \Repeat{
        $\tau \sim D_\pi$ \\ 
        $z_t \leftarrow (f_\theta(\tau_t), f_\theta(\tau_{t+1}))$ \\
        \For{$(x, x^+)^n \ in\ shuffle(z)$}{
             $y_{ij} \leftarrow || x_i - x^+_j ||$ \\
             $L \leftarrow \sum^n_{ki} ReLU(y_{ii} - y_{ik} + \alpha)$\\
             $\theta \leftarrow \theta - \eta \nabla_\theta L $\\
    }{\textbf{until} terminated}}
\caption{S3N Training (triplet loss)}
\label{alg:1}
\end{algorithm}

Part of the difficulty with the approach is in its computational complexity. To avoid computing a distance matrix over an entire trajectory, which is unnecessarily costly, we take a mini-batching approach and employ stochastic gradient descent. Positive pairs $(x_j,x_j^+):= (f_\theta(s_t),f_\theta(s_{t+1}))_j$ are uniformly sampled from trajectories that are collected using a trajectory collector $D_\pi$. For each batch of pairs $(x,x^+)^n$, we assume that the positive part $x_j^+$ for pair $j$ is negative for all other anchors $x_i$ in the batch and construct a distance matrix $y$ accordingly. With a sufficiently large sample space it is unlikely that the assumption is broken, but care should be taken if the graph is dense. In our experiments the effect was negligible. It is also important to note that the embedding dimension should be sufficiently large, with dense graphs requiring larger dimensions. The S3N training algorithm is described in Alg. \ref{alg:1}. Using this algorithm, a good embedding can be learned quickly\footnote{in order of minutes rather than hours using an NVIDIA RTX 2070 GPU} requiring orders of magnitude less data than approaches that rely on prediction or that have a generative aspect.

In order to learn a meaningful embedding, S3N training is semi-supervised and trained only on normal trajectories. In a practical setting, we may not have access to normal trajectories, more likely we have access to an \textit{in-progress} approximate game that contains some bugs. To make S3N viable for use as part of a practical tool, we envisage an active learning procedure in which a developer is continually adapting the training data by re-programming the game after receiving feedback on the most anomalous transitions. As this process continues, the game will approach the ideal game and S3N will improve and adapt its knowledge of normality. Realising active learning is left as future work, in our experiments we use the ideal game directly as an initial demonstration of the feasibility of S3N as an approach.

\begin{figure}
    \centering
    \includegraphics[width=.85\linewidth]{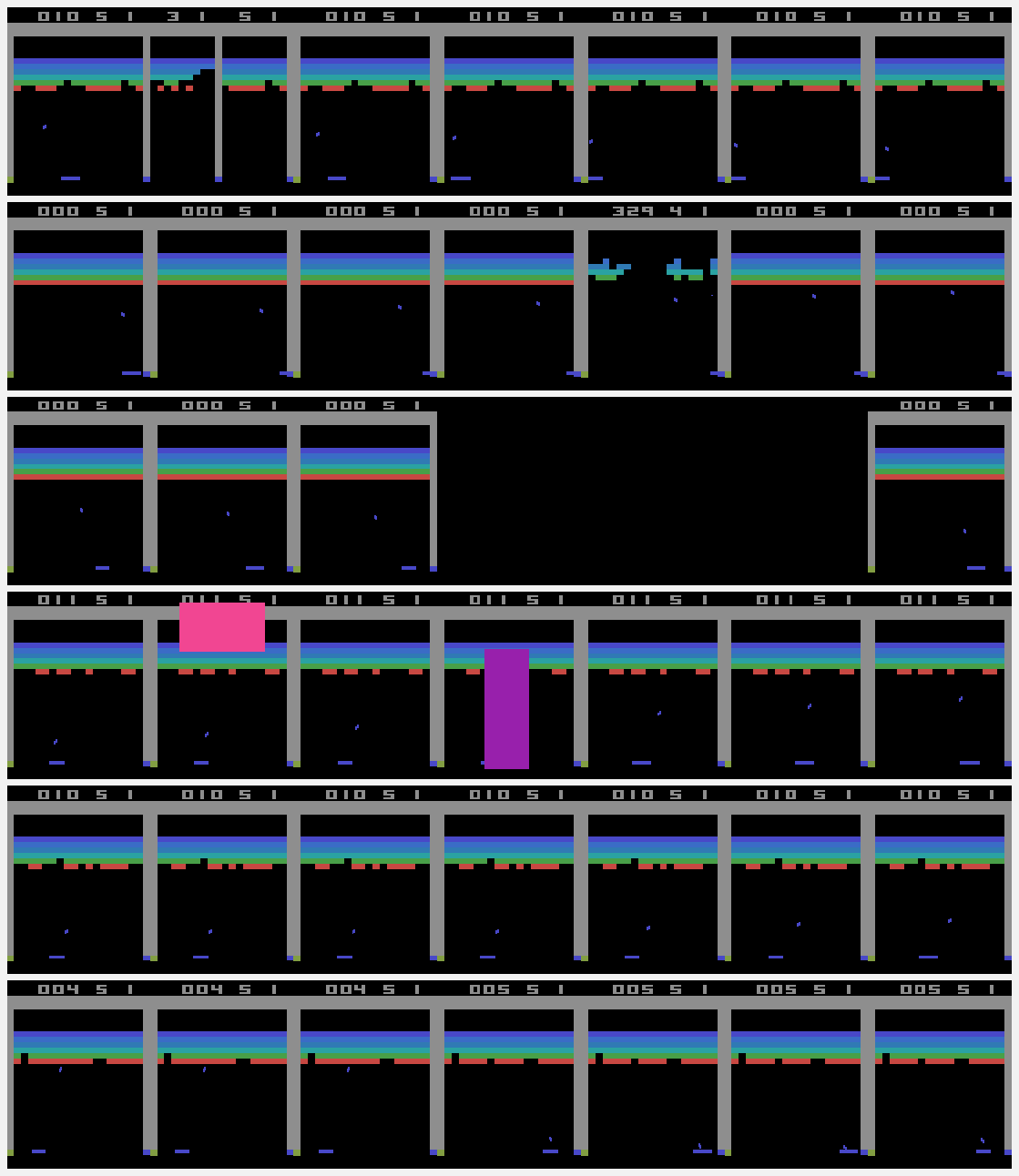}
    \caption{Example anomalies from our dataset (AAD) for Breakout. From top to bottom: split vertical, split horizontal, flickering, visual artefacts, freeze - no frame skip, freeze - frame skip.}
    \label{fig:anomalies}
\end{figure}

\makeatletter
\newcommand{\thickhline}{%
    \noalign {\ifnum 0=`}\fi \hrule height 1pt
    \futurelet \reserved@a \@xhline
}
\newcolumntype{"}{@{\hskip\tabcolsep\vrule width 1pt\hskip\tabcolsep}}
\makeatother

\begin{table*}
    \caption{Table of Results}
    \centering
    {\footnotesize 
    \begin{tabular}{| r " c | c | c | c | c | c " c | c | c | c | c | c |} \hline
     \multicolumn{2}{| c |}{}  & \multicolumn{5}{ c "}{$Pr(\Delta_1 > \Delta_j)$} & \multicolumn{6}{ c |}{AUC$^{\mathrm{1}}$} \\ \hline
     Game (d) & UDS & $j=2$ & $j=5$ & $j=10$ & $j=50$ & $j=100$ & VA & Flicker & Freeze & F-Skip & SH & SV \\ \thickhline
     Beam Rider (64)     & 0.0616 & 0.0517 & 0.0076 & 0.0032 & 0.0019 & 0.0004 & 0.9347 & 0.9997 & 0.0048 & 0.9878 & 0.9905 & 0.9927 \\
     Breakout (256)      & 3.2459 & 0.2403 & 0.0729 & 0.0365 & 0.0028 & 0.0000 & 0.9884 & 1.0000 & 0.0019 & 0.9647 & 0.9791 & 0.9908 \\
     Enduro (256)        & 0.0187 & 0.2197 & 0.0011 & 0.0001 & 0.0000 & 0.0000 & 0.8537 & 1.0000 & 0.0001 & 0.9986 & 0.9828 & 0.9826 \\
     Pong (256)          & 0.0456 & 0.2739 & 0.1093 & 0.0604 & 0.0272 & 0.0156 & 0.9914 & 1.0000 & 0.0044 & 0.9381 & 0.9671 & 0.9697 \\
     Qbert (64)          & 0.0625 & 0.1562 & 0.0185 & 0.0036 & 0.0001 & 0.0000 & 0.9313 & 1.0000 & 0.0048 & 0.9848 & 0.9900 & 0.9820 \\
     Seaquest (64)       & 0.0439 & 0.0969 & 0.0120 & 0.0012 & 0.0000 & 0.0000 & 0.9683 & 1.0000 & 0.0005 & 0.9962 & 0.9929 & 0.9949 \\
     Space Invaders (64) & 0.0284 & 0.0938 & 0.0423 & 0.0245 & 0.0001 & 0.0000 & 0.9834 & 1.0000 & 0.0179 & 0.9750 & 0.9949 & 0.9951 \\ \hline
     \multicolumn{13}{l}{$^{\mathrm{1}}$VA = Visual Artefact, F-Skip = Freeze Skip, SH = Split Horizontal, SV = Split Vertical }
    \end{tabular}}

    \label{fig:resultstable}
\end{table*}

\section{Experiments}
\subsection{Atari Anomaly Dataset}
To test our approach, we use 7 Atari games\footnote{Beam Rider, Breakout, Enduro, Pong, Qbert, Seaquest, and Space Invaders} that have previously been made available as part of the Arcade Learning Environment (ALE) \cite{Bellemare2013} and OpenAI Gym. States (and actions) have been collected using model-free RL, specifically, with the OpenAI stable-baselines \cite{stable-baselines} implementation of Advantage Actor-Critic (A2C)\cite{mnih2016}, totalling approx. 200k states per game. Common types of anomalies \cite{Lewis2010} have been artificially introduced into approximately half of the collected trajectories, these include freezing, flickering and visual artefacts (see Fig. \ref{fig:anomalies}) at a rate of $0.01$. Each game was chosen with a specific motivation in mind, testing S3N's ability to deal with large discontinuities including flashing and scene changes, embed (a)cyclic graphs, dense/sparse graphs, or to deal with a high inherent dimensionality. Data and further details can be found \href{https://www.kaggle.com/benedictwilkinsai/atari-anomaly-dataset-aad}{here}\footnote{https://www.kaggle.com/benedictwilkinsai/atari-anomaly-dataset-aad}.
\newline

\subsection{Implementation Details}
The neural network used in the experiments to follow has a three layer convolutional architecture with leaky ReLU activation and a final linear embedding layer of dimension 64 or 256. The same network architecture was used for each game, with the following set of hyper parameters, batch size $n=128$, margin $\alpha=0.2$, squared $L_2$ norm was used as the distance in triplet loss, learning rate $l=0.0005$ for Adam optimiser. The network was trained for 12 epochs on as little as 60k states from the \textit{raw} partition of AAD. All code and pre-trained models are available \href{https://github.com/BenedictWilkinsAI/S3N}{here}\footnote{\label{note-s3n}https://github.com/BenedictWilkinsAI/S3N}.

\subsection{Results \& Discussion}
Before evaluating the performance of S3N on detecting anomalies, we make an attempt at evaluating the quality of the learned embedding. A poor embedding may be the result of an insufficient embedding dimension or high-entropy transitions, but there are other more subtle possibilities. For example, due to the lack of a hard restriction on the magnitude of $\Delta^t_{t+1}$.

As the learned metric is going to be used directly to determine a ranking for normal and anomalous transitions, in order to avoid false positives, we want to be sure that there are no large jumps in a normal embedding trajectory. At first glance, the standard deviation of displacement $\Delta^t_{t+1}$ seems to give a good indication of uniformity, however self-transitions are an issue. To make the statistic more robust, we look at the standard deviation of the residuals $max(\Delta^t_{t+1} - \alpha, 0)$ where $\alpha$ is the margin parameter. This has the effect of ignoring any normal displacements that are already within an acceptable tolerance, and leads to a more intuitive ideal 0 value. We refer to the standard deviation of residual 1-step displacements as the Uniform Distance Statistic (UDS). 

To ensure the embedding is consistent with the original objective $\Delta^t_{t+1} < \Delta^t_i\ \forall t,i > 1$, we treat each $\Delta_j$ as a random variable whose realisations correspond to $j$-step displacements and determine $Pr(\Delta_1 > \Delta_j | \tau)\ \forall j > 1$ using a rank-sum test. We show results for increasing values of $j$ in Table \ref{fig:resultstable} and see that the probability quickly vanishes. When combined with the UDS, we can conclude that S3N is able to construct good embeddings, even in the face of scene changes and other large discontinuities. In the case of Breakout, UDS is comparatively high. We hypothesise that this is due to its high inherent (combinatorial) dimension with some jumps occurring at the transitions between different block configurations. 

To evaluate the performance of S3N on the detection of anomalies, as is common in the literature, we use the AUC score. As shown by the scores in Table \ref{fig:resultstable}, S3N is able to correctly identify flickering, skips and various kinds of visual artefacts. Freezing is part of a particular class of self-transitioning anomaly that cannot be detected by our approach. In our experiments, S3N is learning a proper distance ($L_2$ norm), i.e. $d(x,y) = d(y,x)$ and $d(x,x) = 0$. The second axiom results in an anomaly score of $0$ being assigned to self-transitions and hence the bad performance in this case. We have given special consideration to labelling transitions for flickering and freeze skip anomalies, labelling only the non self-transitions as anomalous. It should also be noted that S3N is invariant to the direction of time due to symmetry in the distance. We leave it as part of future work to explore alternative measures that might address these issues, perhaps by incorporating action information as a source of asymmetry.

\section{Conclusions \& Future Work}
S3N is an efficient learning algorithm for constructing video game embeddings for the purpose of anomaly detection, requiring orders of magnitude less data and training than similar generative or predictive approaches. We have given an initial demonstration of the feasibility of S3N on our dataset (AAD), making it available to support future work in this area. We have evaluated the ability of S3N to construct meaningful embeddings, and shown that it is able to successfully identify many common types of video game bugs. Future direction includes exploring actions as part of alternative measures for use in the objective.

\bibliographystyle{IEEEtran} \bibliography{IEEEabrv,bib} 

\end{document}